\documentclass[sigconf]{acmart}   
\setcopyright{none}
\acmConference{}{}{}

\acmBooktitle{}
\acmYear{} \acmDOI{} \acmISBN{} \acmPrice{}

\makeatletter
\renewcommand\footnotetextcopyrightpermission[1]{} 
\def\@copyrightpermission{}                        
\makeatother

\AtBeginDocument{%
  }

\usepackage{multirow}

\usepackage{amsmath,amssymb,amsfonts}

\usepackage{algorithm}
\usepackage{algorithmic}

\begin{document}

\title{Larger Hausdorff Dimension in Scanning Pattern Facilitates Mamba-Based Methods in Low-Light Image Enhancement}

\author{Xinhua Wang}
\affiliation{%
  \institution{Imperial College London}
  \city{London}
  \country{UK}}
\email{mw1821@ic.ac.uk}

\author{Caibo Feng}
\affiliation{%
  \institution{University of Sussex}
  \city{Brighton}
  \country{UK}}
\email{2012190104@pop.zjgsu.edu.cn}

\author{Xiangjun Fu}
\affiliation{%
  \institution{University of California San Diego}
  \city{San Diego}
  \country{USA}}
\email{xif001@ucsd.edu}

\author{Chunxiao Liu}
\affiliation{%
  \institution{Zhejiang Gongshang University}
  \city{Hangzhou}
  \country{China}}
\email{cxliu@mail.zjgsu.edu.cn}

\renewcommand{\shortauthors}{Xinhua Wang, Caibo Feng, Xiangjun Fu, Chunxiao Liu}

\begin{abstract}
In the domain of low-light image enhancement, both transformer-based approaches, such as Retinexformer and Mamba-based frameworks, such as MambaLLIE, have demonstrated distinct advantages alongside inherent limitations. Transformer-based methods, in comparison with mamba-based methods, can capture local interactions more effectively, albeit often at a high computational cost. In contrast, Mamba-based techniques provide efficient global information modeling with linear complexity, yet they encounter two significant challenges: (1) inconsistent feature representation at the margins of each scanning row and (2) insufficient capture of fine-grained local interactions. To overcome these challenges, we propose an innovative enhancement to the Mamba framework by increasing the Hausdorff dimension of its scanning pattern through a novel Hilbert Selective Scan mechanism. This mechanism explores the feature space more effectively, capturing intricate fine-scale details and improving overall coverage. As a result, it mitigates information inconsistencies while refining spatial locality to better capture subtle local interactions without sacrificing the model’s ability to handle long-range dependencies. Extensive experiments on publicly available benchmarks demonstrate that our approach significantly improves both the quantitative metrics and qualitative visual fidelity of existing Mamba-based low-light image enhancement methods, all while reducing computational resource consumption and shortening inference time. We believe that this refined strategy not only advances the state-of-the-art in low-light image enhancement but also holds promise for broader applications in fields that leverage Mamba-based techniques. Our code is available \textbf{\textit{\href{https://github.com/Murpphhyy/HausdorffMamba}{here}}.}
\end{abstract}

\maketitle

\section{Introduction}

Low-light image enhancement is an essential but challenging topic in computer vision. It aims to restore the image degradation with low brightness, improve the visual quality of images captured under low-light conditions, and aid high-level visual tasks (e.g., object detection, face recognition, and action detection). 

Numerous traditional methods, such as histogram equalization \cite{cheng2004simple, huang2012efficient} and Retinex theory \cite{jobson1997multiscale, gu2019novel}, have been proposed to enhance low-light images. With the development of deep learning, many learning-based methods \cite{wei2018deep,zhang2019kindling,li2021low,liu2021retinex,zhang2021beyond,zamir2022learning,ma2022toward,cai2023retinexformer} have emerged, which improve visual quality by learning the mapping between low-light and normal-light images using end-to-end models or deep Retinex-based models to make them more robust than traditional approaches. Recently, some methods \cite{xu2022illumination,jiang2023low} have explored wavelet transformation for low-light image enhancement, integrating wavelet information into convolutional neural networks (CNNs) or transformer models to achieve impressive results. However, CNNs \cite{zhang2021beyond} have limitations in modeling long-range dependencies and non-local self-similarity, resulting in challenges in addressing image degradation effectively in low-light scenarios. In addition, the quadratic growth of the complexity of transformer models \cite{cai2023retinexformer} results in inefficient use of computational resources. Both the CNN model \cite{guo2023low} and the transformer model \cite{xu2022snr} achieve average performance only in both aspects of performance, as indicated by the PSNR scores and inference time.

To overcome these issues, Mamba \cite{gu2023mamba} is designed to model long-range dependencies and enhance the efficiency of training and inference through a selection mechanism and a hardware-aware algorithm. For now, numerous studies have explored the applications of Mamba in computer vision. Vision Mamba \cite{vim} introduces the Vim block, incorporating a bidirectional state space model for efficient learning. Meanwhile, VMamba \cite{liu2024vmamba} introduces a CSM module to traverse the spatial domain, combining the advantages of CNNs and Visual Transformers (ViTs) while achieving computational efficiency in linear complexity without sacrificing the global receptive field. Additionally, U-Mamba \cite{ma2024u} combines U-Net and Mamba models to enhance medical image segmentation. SegMamba \cite{xing2024segm} utilises Mamba's efficient reasoning and linear scalability to enable fast and accurate processing of large-scale 3D medical data. Additionally, WaveMamba \cite{zou2024wavemamba}, RetinexMamba \cite{bai2024retinexmamba}, and MambaLLIE \cite{MambaLLIE} demonstrate marvelous performance of Mamba in the field of low-light image enhancement, all of which generally applies Mamba along with U-Net backbone, significantly reducing the computational complexity compared with Retinexformer \cite{cai2023retinexformer}.

However, several challenges arise when incorporating Mamba with vision tasks. The primary challenge originates from the mismatch between the causal sequential modeling of Mamba and the two-dimensional (2D) data structure of images. Mamba is designed for one-dimensional (1D) causal modeling for sequential signals, which cannot be directly leveraged for modeling two-dimensional image tokens. A simple solution is to use the raster-scan order to convert 2D data into 1D sequences. However, it restricts the receptive field of each location to only the previous locations in the raster-scan order, which means such type of raster scan, in some cases, fails to capture local interactions when dealing with features with more extensive areas in the images. Moreover, in the raster-scan order, the ending of the current row is followed by the beginning of the next row, while they do not share spatial continuity. 

To accurately evaluate the ability of a scanning path to capture complicated patterns in an image, we propose the Hausdorff Dimension Measurement Method. The Hausdorff dimension of a scanning curve directly reflects its inherent complexity and capacity to capture local interactions. A higher Hausdorff dimension indicates that the curve is not simply a smooth, predictable path but rich in intricate detail and subtle fluctuations. This complexity enables the scanning pattern to more effectively perceive local variations, ensuring that fine-scale interactions within the scanned area are not overlooked. In other words, as the Hausdorff dimension increases, it signals a more nuanced and densely packed trajectory that can more effectively capture local features while maintaining the ability of Mamba to understand long-range dependencies.

Besides, we conduct extensive experiments to show that Mamba scanning patterns with higher Hausdorff dimension, Hilbert scan, and Peano scan, inspired by Hilbert curve \cite{hilbert1891} and Peano curve \cite{peano1890sur}, can effectively prompt the performance of previous Mamba-based low-light image enhancement methods on several paired datasets. Hilbert and Peano scans not only leverage the Hausdorff Dimension of 2 but also map neighboring 2D points to nearby positions in the 1D sequence, which minimizes discontinuities and improves cache performance and data coherence. The visual presentations of different scanning patterns are provided in the supplementary material.

Our main contributions are summarized as follows:
\begin{itemize}
\item We propose an innovative method for evaluating scanning paths by employing the Hausdorff dimension. This metric measures the inherent complexity of a scanning curve, thereby assessing its capacity to capture intricate, fine-scale local variations and to effectively navigate complex image structures.
\item We proved that scanning patterns with high Hausdorff Dimension (eg, Hilbert scan, and Peano scan) leverage the curve’s intrinsic properties to capture local features effectively in larger regions, mapping neighboring 2D points to adjacent positions in the 1D sequence.
\item Experimental results on a comprehensive set of benchmark datasets consistently demonstrate the reliability of the measure of Hausdorff Dimension and the effectiveness of Hilbert scan and Peano scan on improving the performance of existing Mamba-based low-light image enhancement methods. 
\end{itemize}

\section{Related Work}

\subsection{Low-light Image Enhancement}
Following the development of learning-based image restoration methods \cite{wang2018gladnet, yang2020fidelity, lim2020dslr, liang2022semantically}, LLNet \cite{lore2017llnet} synthesizes paired data by applying gamma adjustment and randomly adding noise to clean images. MBLLEN \cite{lv2018mbllen} extracts features at different levels using a multi-scale network structure to achieve better results. LightenNet \cite{li2018lightennet} directly estimates the illumination map based on the input. RetinexNet \cite{wei2018deep} utilizes the Retinex theory to decompose low-light images into reflection and illumination components. ZeroDCE \cite{guo2020zero} proposed to transform LLIE into a curve estimation problem and designed a zero-reference learning strategy for training. EnlightenGAN \cite{jiang2021enlightengan} adopted unpaired images for training for the first time by using the generative adverse network as the main framework. KinD++ \cite{zhang2021beyond} utilizes a layer-wise decomposition strategy for brightness adjustment and detail reconstruction to suppress noise in the reflection layer further. LLFlow \cite{wang2022low} uses conditional normalization flows for low-light image enhancement. Restormer \cite{zamir2022restormer} designs a lightweight transformer for image restoration tasks. SNRNet \cite{xu2022snr} introduces an SNR-aware CNN-transformer hybrid network for low-light image enhancement. MBPNet \cite{zhang2023multi} introduces a multi-branch progressive network for low-light image enhancement. Bread \cite{guo2023low} decomposes low-light images into texture and chrominance components and suppresses complex noise through adjustable noise suppression networks. Retinexformer \cite{cai2023retinexformer} combined the Retinex theory with the design of a one-stage transformer, further refining and optimizing this approach. Diff-Retinex \cite{Diff-Retinex} designed a transformer-based decomposition network and adopted generative diffusion networks to reconstruct the results. Overall, they typically applied the Retinex theory directly, which may be limited for low light enhancement problems.

\subsection{Mamba-based Methods}
Recently, the success of State Space Models (SSMs) \cite{gu2021efficiently} has been increasingly recognized as a promising direction in research, which is proposed as a novel alternative to CNNs or Transformers to model long-range dependency. Contemporary SSMs such as Mamba \cite{gu2023mamba} not only establish long-distance dependency relations but also demonstrate linear complexity with respect to input size. Other methods introduce the state space models for visual applications. UMamba \cite{ma2024u} proposes a hybrid CNN-SSM architecture to handle the long-range dependencies in biomedical image segmentation. Vision Mamba \cite{vim} suggests that pure SSM models can serve as a generic visual backbone. SegMamba \cite{xing2024segm} introduces a novel 3D medical image segmentation model designed to capture long-range dependencies within volume features at every scale effectively. Swin-UMamba \cite{liu2024swin} introduces a medical image segmentation model based on Mamba, which utilizes pre-trained models from ImageNet to improve the performance of medical image segmentation tasks. Furthermore, LocalMamba \cite{LocalMamba} was focused on the local scanning strategy and preservation of local context dependencies. EfficientVMamba \cite{EfficientVMamba} designed a lightweight SSMs with an additional convolution branch to learn both global and local representational features. MambaIR \cite{MambaIR} employed convolution and channel attention to enhance the capabilities of the Mamba. WaveMamba \cite{zou2024wavemamba} focuses on applying Wavelet Transform alongside Mamba to mitigate information loss and address the limitation of SSMs, which struggle to model noise effectively. In the case of low-light image enhancement, this insensitivity could lead to the inability to detect or leverage subtle noise patterns that carry important information effectively. RetinexMamba \cite{bai2024retinexmamba} uses SS2D to replace Transformers in capturing long-range dependencies. MambaLLIE \cite{MambaLLIE} introduces a novel global-then-local state space block that integrates a local-enhanced state space module and an implicit Retinex-aware selective kernel module. This design effectively captures intricate global and local dependencies. However, although a number of applications of the Mamba-based image restoration approach have been proposed, none of them have researched the factors determining the effectiveness of a scanning pattern of Mamba.

\setlength{\parindent}{15pt}
\setlength{\parskip}{0pt}

\section{Preliminaries}

\subsection{State Space Model}

SSMs, such as the structured state space sequence models (S4)~\cite{gu2021efficiently} and Mamba~\cite{gu2023mamba}, can be interpreted as continuous linear time-invariant (LTI) systems~\cite{LTI}. Given a one-dimensional input sequence \(x(t) \in \mathbb{R}\), these models map it to an output sequence \(y(t) \in \mathbb{R}\) by means of a hidden state \(h(t) \in \mathbb{R}^m\), where \(m\) is the dimension of the hidden state. The entire system is governed by the following linear ordinary differential equations:
\begin{align}
\begin{aligned}
h'(t) &= {\bf A}\,h(t) + {\bf B}\,x(t), \\
y(t)  &= {\bf C}\,h(t) + {\bf D}\,x(t).
\end{aligned}
\end{align}
Here, \({\bf A} \in \mathbb{R}^{m \times m}\) is the state matrix, \({\bf B} \in \mathbb{R}^{m \times 1}\) represents the input projection, \({\bf C} \in \mathbb{R}^{1 \times m}\) is the output projection, and \({\bf D} \in \mathbb{R}\) denotes the feedthrough parameter.

Since these state-space models are inherently continuous, they must be discretized for implementation on a computer. Using the zero-order hold (ZOH) method, the continuous matrices \({\bf A}\) and \({\bf B}\) are converted into their discrete counterparts \({\bf \bar{A}}\) and \({\bf \bar{B}}\) as follows:
\begin{align}
\begin{array}{l}
{\bf \bar{A}} = \exp\left(\Delta\,{\bf A}\right),\quad 
{\bf \bar{B}} = \left(\Delta\,{\bf A}\right)^{-1}\left(\exp\left(\Delta\,{\bf A}\right) - {\bf I}\right) \cdot \Delta\,{\bf B},
\end{array}
\end{align}
where \(\Delta\) is the step size. Thus, the discrete formulation becomes:
\begin{align}
\begin{array}{l}
h_t = {\bf \bar{A}}\,h_{t-1} + {\bf \bar{B}}\,x_t,\quad
y_t = {\bf C}\,h_t + {\bf D}\,x_t.
\end{array}
\end{align}

However, this formulation remains static with respect to varying inputs. To overcome this limitation, Mamba~\cite{gu2023mamba} introduces selective state-space models, where the parameters dynamically adapt based on the input, which is formulated as:
\begin{equation}
\begin{aligned}
\overline{{\bf B}} &= f_{{\bf B}}(x_t), \quad 
\overline{{\bf C}} = f_{{\bf C}}(x_t), \quad 
\Delta = \vartheta_{{\bf A}}\Bigl({\bf P} + f_{{\bf A}}(x_t)\Bigr),
\end{aligned}
\end{equation}
with \(f_{{\bf B}}(x_t)\), \(f_{{\bf C}}(x_t)\), and \(f_{{\bf A}}(x_t)\) being linear functions that expand the input features into the hidden state space. Although SSMs are effective at modeling long sequences, they may struggle to capture complex local details. To address this challenge in visual data, methods such as VMamba~\cite{liu2024vmamba} and Vim~\cite{vim} employ specialized location-aware scanning strategies that preserve the two-dimensional structure of images.

\subsection{Hausdorff Dimension}
The Hausdorff dimension provides a robust and precise method of quantifying the complexity of geometric objects, including fractals, whose dimensions are often non-integer and, therefore, cannot be described by traditional geometric measures \cite{mandelbrot1982fractal, edgar2007measure}.

To understand Hausdorff Dimension, we must first introduce the definition of Hausdorff Measure. Let \( S \subseteq \mathbb{R}^n \) be a subset, a geometric object. For each real number \( d \geq 0 \) and every \( \epsilon > 0 \), define the \( d \)-dimensional Hausdorff content \(H^d_{\epsilon}(S)\) by:
\begin{equation}
\begin{aligned}
H^d_{\epsilon}(S) = \inf \left\{ \sum_{i=1}^{\infty} (\text{diam}(U_i))^d : S \subseteq \bigcup_{i=1}^{\infty}U_i,\, \text{diam}(U_i)\leq \epsilon \right\}.
\label{HDmeasure}
\end{aligned}
\end{equation}
where \(\text{diam}(U_i)\) is defined as \(\text{diam}(U_i) = \sup\{|x-y| : x,y\in U_i\}\) represents the greatest distance between any two points within within subset \(U_i\). \(U_i\) are subsets that "cover" the set \(S\). The set \(S\) must be fully contained within the union \(\bigcup_{i=1}^{\infty}U_i\). Typically, these can be intervals, balls, or cubes. \(\epsilon\) is a positive real number representing the scale we measure. Smaller \(\epsilon\) means subsets \(U_i\) have smaller sizes.  The infimum is then taken over all countable coverings \(\{U_i\}\) of \(S\) \cite{edgar2007measure} as a measure of the total size of set \(S\).

Then the \( d \)-dimensional Hausdorff measure \(H^d(S)\) is defined as:
\begin{equation}
\begin{aligned}
H^d(S)=\lim_{\epsilon\rightarrow 0}H_{\epsilon}^{d}(S).
\end{aligned}
\end{equation}

Given the Hausdorff measure, the Hausdorff dimension \( \dim_H(S) \) of the set \( S \subseteq \mathbb{R}^n \) is defined as:
\begin{equation}
\begin{aligned}
\dim_H(S)=\inf\{d : H^{d}(S)=0\} = \sup\{ d : H^d(S)=\infty \}.
\end{aligned}
\end{equation}

Referring to equation \ref{HDmeasure}, as \(\epsilon\rightarrow 0\), the value of \(\text{diam}(U_i)\) will be very small, which means when \(d\) becomes too large, the value of \(H^{d}(S)\) would become 0. Vice versa, when \(d\) becomes too small, the value of \(H^{d}(S)\) becomes infinite. Intuitively, this dimension identifies a critical threshold separating two regimes: dimensions for which the measure is infinite and dimensions for which it collapses to zero \cite{edgar2007measure}. 

The concept of Hausdorff dimension has wide-ranging implications across mathematics and applied fields, particularly in analyzing the complexity and scaling properties of fractal objects, chaotic dynamics, and geometric complexity in nature and data sciences \cite{mandelbrot1982fractal, edgar2007measure}.

\section{Methodology}
In this section, we first introduce how the Hausdorff Dimension of a scanning pattern of Mamba influences Mamba-based low-light image enhancement methods and then move on to the explanation of the superiority of the Hilbert scan and Peano scan in Mamba-based low-light enhancement methods.

\subsection{Hausdorff Dimension's Implications for Vision Mamba}
In many vision applications, image processing and feature extraction depend on how well a discrete set of samples approximates a continuous image function. In particular, the scanning pattern used to traverse the spatial domain of an image can have significant implications for the performance of downstream tasks. This section investigates the effect of employing a scanning pattern with a high Hausdorff dimension. This intuitively leads to more uniform and space-filling coverage, thus reducing the worst-case approximation error.

\noindent \textbf{Spatial Domain \& Sampling Set: } We define $\Omega$ as the spatial domain of the original input images:
\[
\Omega \subset \mathbb{R}^n,
\]
where $n$ denotes the number of spatial dimensions. For example, for a two-dimensional image of height $H$ and width $W$, we have
\[
\Omega = \{(x, y) \mid 0 \le x < H,\; 0 \le y < W\}.
\]

It should be noted that $\Omega$ represents the set of all spatial coordinates on which the image is defined. The image itself can be described as a function.
\[
f : \Omega \to \mathbb{R}^c,
\]
where $c$ is the number of channels, e.g., $c=3$ for an RGB image. Note that $\Omega$ does not include any feature content of the image; it merely defines the locations over which the image $f$ is defined.

We define $P$ as the sampling set of the spatial domain $\Omega$:
\[
P \subset \Omega,
\]
where $P$ is a collection of points from which the scanning algorithm extracts data. While a full ordinary raster scan in Mamba might have $P = \Omega$ in a discrete sense, advanced architectures such as Vision Mamba often employ selective or non-uniform scanning strategies. In such cases, $P$ is a proper subset of $\Omega$, and the distribution of points in $P$, their density, uniformity, and fractal properties, directly influence the quality of the approximation of the continuous function $f$, which represents the original image.

\noindent \textbf{Function Spaces and Smoothness Conditions: } To rigorously analyze how well the continuous image function $f : \Omega \to \mathbb{R}^c$ is approximated by its samples, we place $f$ within appropriate function spaces and assume certain smoothness conditions. These conditions, such as $L^2$ integrability, Lipschitz continuity, and H\"older continuity, provide a mathematical framework that allows us to derive error bounds on the approximation quality. In this subsection, we describe these spaces and conditions in detail.

A function $f$ belongs to $L^2(\Omega)$ if 
\begin{equation}
\|f\|_{L^2(\Omega)} = \left(\int_\Omega |f(x)|^2\,dx\right)^{1/2} < \infty.
\end{equation}

This condition ensures that $f$ has finite energy, a natural requirement in signal processing and machine learning.

A function $f: \Omega \to \mathbb{R}$ is said to be Lipschitz continuous if there exists a constant $L_f > 0$ such that
\begin{equation}
|f(x) - f(y)| \le L_f\,\|x-y\| \quad \forall\, x,y\in\Omega.
\end{equation}

This condition bounds the rate at which $f$ can change between any two points in $\Omega$, ensuring that the function does not exhibit abrupt transitions.

More generally, $f$ is H\"older continuous with exponent $\alpha\in (0,1]$ if there exists a constant $C_f>0$ such that
\begin{equation}
|f(x) - f(y)| \le C_f\,\|x-y\|^{\alpha} \quad \forall\, x,y\in\Omega.
\end{equation}

When $\alpha=1$, H\"older continuity is equivalent to Lipschitz continuity. For $\alpha < 1$, the condition allows for more gradual variations in $f$. These smoothness properties are essential for establishing rigorous error estimates in function approximation.

\noindent \textbf{Worst-Case Approximation Error from Sparse Sampling: }
This subsection examines the error incurred when approximating a continuous function from discrete samples, which is central to understanding the performance impact of different scanning patterns.

\noindent \textbf{Dispersion \& Worst-Case Approximation Error: } 
The \emph{dispersion} of a sampling set $P$ in $\Omega$ is defined as
\begin{equation}
\varepsilon(P,\Omega) = \sup_{x\in\Omega} \min_{p\in P} \|x-p\|.
\end{equation}

For each point $x\in \Omega$, the term $\min_{p\in P} \|x-p\|$ is the distance from $x$ to its nearest sample in $P$. Taking the supremum over $x\in \Omega$ yields the largest distance, quantifying the worst-case gap in the sampling coverage. A smaller $\varepsilon(P,\Omega)$ indicates that every point in $\Omega$ is close to at least one sample, which is critical for accurate approximation.

Let $\mathcal{F}$ denote the class of functions defined on $\Omega$ that belong to $L^2(\Omega)$ and satisfy Lipschitz or H\"older continuity. When approximating a function $f\in\mathcal{F}$ using only its values at points in $P$, interpolation theory provides a pointwise error bound:
\begin{equation}
|f(x)-\hat{f}(x)| \le K\,\varepsilon(P,\Omega)^\alpha,\quad \forall\, x\in\Omega,
\label{errorbound}
\end{equation}
where $\hat{f}(x)$ is an interpolant of $f$ from the samples \(P\), and $K$ is a constant dependent on the smoothness constants (e.g., $C_f$ or $L_f$) \cite{devore1993constructive}, and $\alpha$ is the H\"older exponent. Integrating this error over $\Omega$ in the $L^2$ norm yields the worst-case approximation error:
\begin{equation}
E(P,\mathcal{F}) = \sup_{f\in\mathcal{F}} \inf_{\hat{f}} \|f-\hat{f}\|_{L^2(\Omega)} \le K'\,\varepsilon(P,\Omega)^\alpha.
\end{equation}

This relation indicates that minimizing the dispersion $\varepsilon(P,\Omega)$ is key to reducing the overall approximation error 
 in any $f\in\mathcal{F}$ \cite{devore1993constructive}. If the sampling points $P$ are sparse or irregularly distributed, $\varepsilon(P,\Omega)$ will be larger, leading to a higher worst-case error. Hence, having a scanning pattern that minimizes this worst-case error is critical for faithful approximation.

\noindent \textbf{Scanning Patterns \& Hausdorff Dimension: } In this subsection, we discuss how the fractal properties of a scanning pattern, as measured by its Hausdorff dimension, impact the dispersion and, consequently, the approximation error.

The Hausdorff dimension $\dim_H(P)$ of a set $P$ is defined using the $s$-dimensional Hausdorff measure:
\begin{equation}
\mathcal{H}^s(P) = \inf \left\{ \sum_{i=1}^{\infty} (\text{diam}(U_i))^s : P \subseteq \bigcup_{i=1}^{\infty}U_i,\, \text{diam}(U_i)\leq \delta \right\}.
\end{equation}
where \(\delta \to 0\), and
\[
\dim_H(P) = \inf\{ s : \mathcal{H}^s(P)=0\}.
\]

A higher Hausdorff dimension implies that $P$ is more space-filling and provides a more uniform coverage of $\Omega$. This concept has been thoroughly discussed in the literature \cite{mandelbrot1982fractal, falconer2003fractal} and plays a key role in determining the uniformity of the sampling.

Under reasonable uniformity assumptions, if two sampling sets $P_1$ and $P_2$ satisfy
\[
\dim_H(P_2)>\dim_H(P_1),
\]
Then generally, according to \cite{mandelbrot1982fractal, devore1993constructive, falconer2003fractal}, as scanning patterns with higher Hausdorff Dimensions cover the continuous image function $f$ more uniformly, higher Hausdorff Dimensions lead to lower dispersion value:
\[
\varepsilon(P_2,\Omega) \le \varepsilon(P_1,\Omega).
\]
which directly leads to a lower worst-case approximation error.

Consider two scanning methods: ordinary raster scan and Hilbert scan.
Raster scan follows a simple row-by-row order. Its continuous parameterization is effectively one-dimensional (Hausdorff dimension of one) even though it visits all discrete points.
In contrast, Hilbert scan is a space-filling curve with Hausdorff dimension of two \cite{mandelbrot1982fractal, falconer2003fractal}, designed to preserve more input features with its fractal nature.

While both methods may visit every pixel in a discrete grid, in scenarios of selective scan in Mamba, the Hilbert curve typically yields a lower dispersion, thereby providing a more robust reconstruction of the underlying image function \cite{mandelbrot1982fractal, falconer2003fractal, matousek1999geometric}. The same principle can also be applied to evaluate the effectiveness of the Peano scan.

\begin{table*}[t]
\setlength{\abovecaptionskip}{0.1cm} 
\setlength{\belowcaptionskip}{-0.3cm}
\centering
\caption{Quantitative comparisons of different methods on LOLv1 \cite{wei2018deep}, LOLv2-real \cite{yang2021sparse} and LOLv2-synthetic \cite{yang2021sparse}. The best and second-best results are highlighted in \textbf{bold} and \underline{underlined}, respectively. Note that we download the pre-trained models from the authors' websites.}
\begin{tabular}{c|ccc|ccc|ccc}
\toprule
\multicolumn{1}{c|}{\multirow{2}{*}{Methods}} &
\multicolumn{3}{c|}{LOLv1} & 
\multicolumn{3}{c|}{LOLv2-real} &
\multicolumn{3}{c}{LOLv2-synthetic}
\\
\cmidrule(r){2-4} \cmidrule(r){5-7} \cmidrule(r){8-10} 
& PSNR$\uparrow$ & SSIM$\uparrow$ & LPIPS$\downarrow$
& PSNR$\uparrow$ & SSIM$\uparrow$ & LPIPS$\downarrow$
& PSNR$\uparrow$ & SSIM$\uparrow$ & LPIPS$\downarrow$
\\ \midrule 

RetinexNet \cite{wei2018deep}
&16.77 &0.419 &0.376
&16.10 &0.401 &0.437
&17.14 &0.761 &0.204
\\ 

MBLLEN \cite{lv2018mbllen}
&17.86 &0.727 &0.153
&17.78 &0.694 &0.193
&16.10 &0.696 &0.195
\\ 

ZeroDCE \cite{guo2020zero}
&14.86 &0.559 &0.237
&18.06 &0.574 &0.216
&17.76 &0.816 &0.126
\\ 

MIRNet \cite{zamir2020learning}
&24.14 &0.840 &0.093
&20.02 &0.820 &0.233
&15.76 &0.735 &0.189
\\ 

EnlightenGAN \cite{jiang2021enlightengan}
&17.48 &0.651 &0.226
&18.64 &0.675 &0.219
&16.57 &0.775 &0.170
\\ 

KinD++ \cite{zhang2021beyond}
&21.80 &0.834 &0.108
&22.21 &0.843 &0.122
&19.26 &0.806 &0.180
\\ 

LLFlow \cite{wang2022low}
&19.34 &0.840 &0.095
&24.15 &0.864 &\textbf{0.060}
&16.89 &0.801 &0.166
\\ 

URetinexNet \cite{wu2022uretinex}
&20.14 &0.823 &0.089
&19.78 &0.843 &0.088
&18.77 &0.824 &0.141
\\ 

SNRNet \cite{xu2022snr}
&\underline{24.61} &0.842 &0.107
&21.48 &0.849 &0.109
&24.14 &0.908 &0.090
\\ 

Bread \cite{guo2023low}
&20.62 &0.834 &0.108
&23.69 &0.861 &0.101
&15.97 &0.748 &0.204
\\ 

LANet \cite{yang2023learning}
&21.74 &0.820 &0.101
&25.30 &0.859 &0.095
&16.99 &0.743 &0.241
\\ 

FourLLIE \cite{wang2023fourllie}
&20.03 &0.820 &0.088
&22.35 &0.847 &\underline{0.071}
&24.65 &0.910 &0.047
\\ 

Retinexformer \cite{cai2023retinexformer}
&23.50 &0.831 &0.092
&22.79 &0.840 &0.110
&\underline{25.39} &0.929 &\underline{0.042}
\\ 

RetinexMamba \cite{bai2024retinexmamba}
&23.34 &0.833 &0.089
&22.45 &0.844 &0.118
&25.31 &0.9291 &0.041
\\ 

MambaLLIE \cite{MambaLLIE}
&21.29 &0.824 &0.091
&22.95 &0.847 &0.105
&24.61 &0.9293 &0.043
\\ 

WaveMamba \cite{zou2024wavemamba}
&23.01 &0.835 &0.1325
&29.04 &0.908 &0.089
&24.63 &0.923 &0.074
\\ 

PeanoMamba (Ours)
&23.99 &\underline{0.857} &0.101
&\underline{30.71} &\textbf{0.918} &0.0778
&25.32 &\underline{0.938} &0.046
\\ 

HilbertMamba (Ours)
&\textbf{24.85} &\textbf{0.862} &\textbf{0.117}
&\textbf{31.04} &\underline{0.915} &0.086
&\textbf{25.74} &\textbf{0.934} &\textbf{0.050}
\\ \bottomrule
\end{tabular}
\label{tab1}
\end{table*}

\begin{figure}[t]
\setlength{\abovecaptionskip}{0.01cm} 
\setlength{\belowcaptionskip}{-0.3cm} 
  \centering
  \includegraphics[width=1\linewidth]{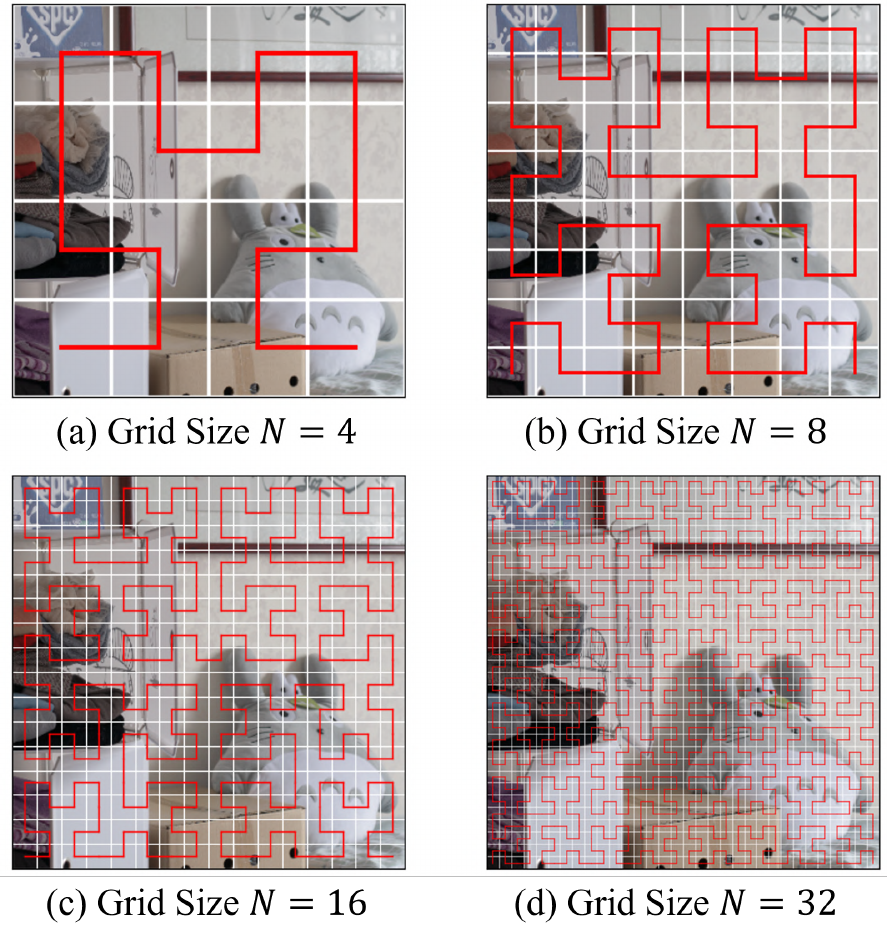}
  \caption{Visualization of self-similar nature of Hilbert curve \cite{hilbert1891}. Proper rotation and duplication can transform the pattern in (a) into (b), the second-order Hilbert curve. Vice versa for (c) and (d).}
  \label{fig:hilbertscan}
\end{figure}

\subsection{Hilbert Scan \& Peano Scan}

In this subsection, we introduce two distinct low-light enhancement methods, HilbertMamba and PeanoMamba, that leverage the superiority of the Hilbert curve \cite{hilbert1891} and Peano curve \cite{peano1890sur}. The Hilbert Scan builds on the principles of space-filling curves. It leverages the fractal geometry of the Hilbert curve and Peano curve \cite{hilbert1891, peano1890sur, mandelbrot1982fractal} to address the challenges associated with mapping two-dimensional image data to a one-dimensional sequence. Both methods are based on WaveMamba \cite{zou2024wavemamba}, where we replace its raster-scanning strategy with Hilber-scanning and Peano-scanning strategy, respectively.

\noindent \textbf{HilbertMamba:}  
Based on WaveMamba \cite{zou2024wavemamba}, HilbertMamba leverages the superiority of the Hilbert scan, a scanning strategy inspired by the Hilbert curve \cite{hilbert1891}, which is defined as a continuous mapping:
\begin{equation}
H: [0,1] \rightarrow [0,1]^2,
\end{equation}
that fills the unit square. In our context, the Hilbert Scan reorders the pixels of an \(H \times W\) image grid into a one-dimensional sequence, denoted by \(I(x,y)\), such that adjacent 2D coordinates are mapped to nearby positions in the sequence. As shown in Figure \ref{fig:hilbertscan}, the Hilbert curve has a self-similar nature, which means the Hilbert curve at higher orders can always be formed by proper rotation and duplication of patterns of lower orders. Formally, for a pixel at spatial location \((x,y) \in \Omega\), the Hilbert Scan assigns an index: 
\begin{equation}
I(x,y) = H^{-1}(x,y),
\end{equation}
thereby preserving the continuity of the spatial domain and ensuring that local neighborhoods remain coherent after transformation.

\begin{figure}[t]
\setlength{\abovecaptionskip}{0.01cm} 
\setlength{\belowcaptionskip}{-0.3cm} 
  \centering
  \includegraphics[width=1\linewidth]{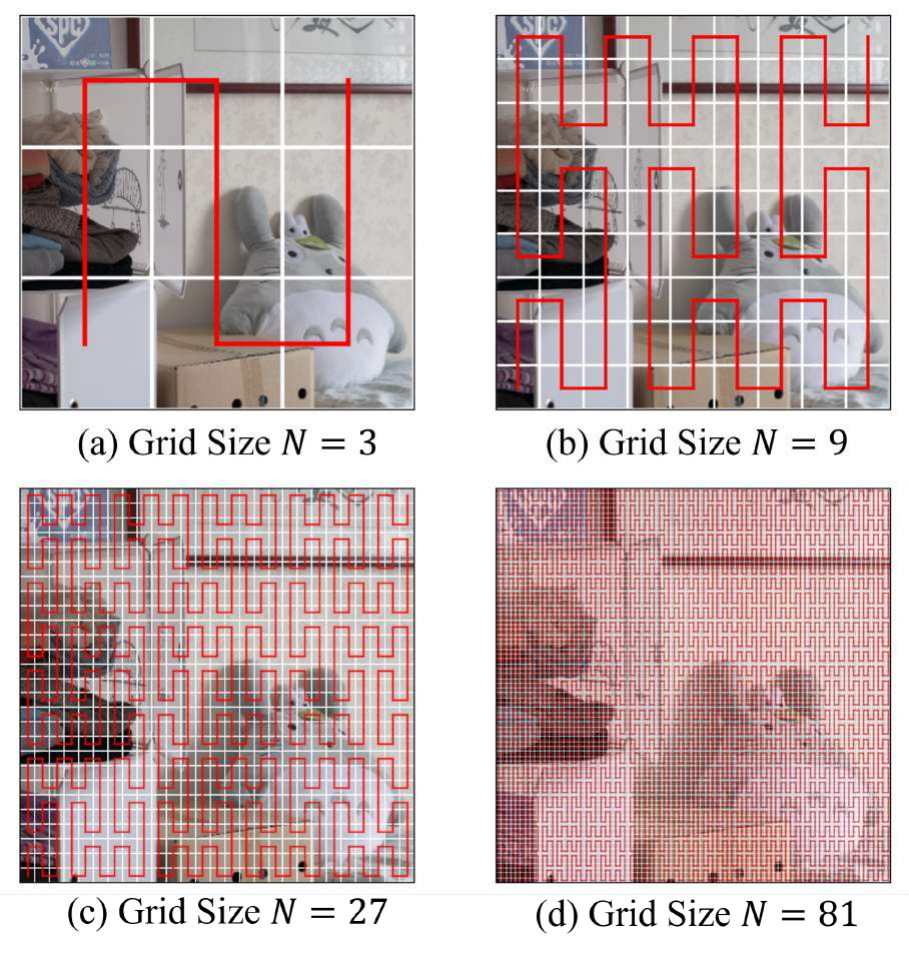}
  \caption{Visualization of self-similar nature of Peano curve \cite{peano1890sur}. Proper rotation and duplication can transform the pattern in (a) into (b), the second-order Peano curve. Vice versa for (c) and (d).}
  \label{fig:peanoscan}
\end{figure}

\noindent \textbf{PeanoMamba:}
PeanoMamba utilizes the Peano scan, inspired by the Peano curve \cite{peano1890sur}, another space-filling curve, offering a distinct fractal approach compared to the Hilbert curve. The Peano curve similarly provides a continuous mapping:
\begin{equation}
P: [0,1] \rightarrow [0,1]^2,
\end{equation}
and ensures comprehensive coverage of the image domain through a unique traversal pattern. The Peano scanning patterns with different grid sizes are demonstrated in \ref{fig:peanoscan}. Analogous to HilbertMamba, PeanoMamba transforms a two-dimensional  image into a one-dimensional sequence, preserving local spatial coherence through the indexing:
\begin{equation}
I(x,y) = P^{-1}(x,y).
\end{equation}
Peano scan's distinct fractal structure effectively maintains both fine-grained local details and global dependencies, contributing uniquely to low-light image enhancement.

\begin{figure*}[t]
\setlength{\abovecaptionskip}{0.01cm} 
\setlength{\belowcaptionskip}{-0.3cm} 
\centering
\includegraphics[width=1\linewidth]{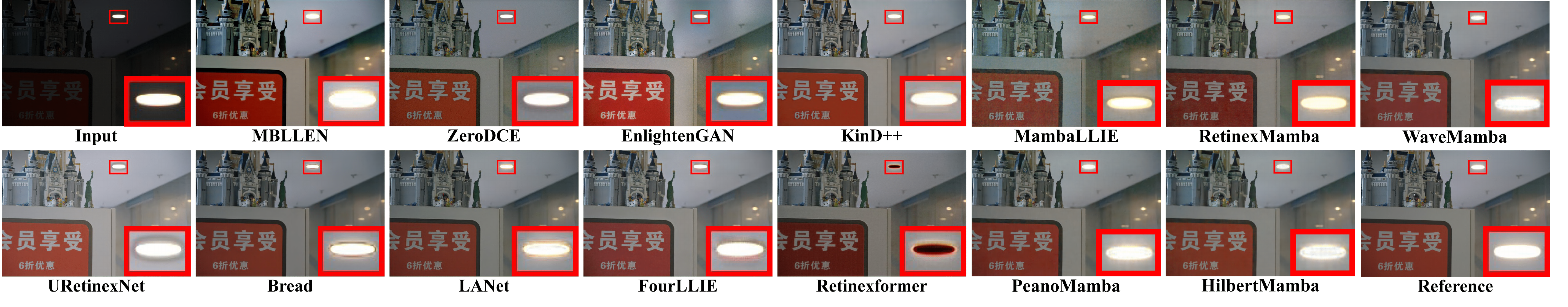}
\caption{\label{lolv2_real}%
Visual comparisons of the enhanced results by different methods on LOLv2-real.}
\end{figure*}

\noindent \textbf{Superiority of HilbertMamba \& PeanoMamba:}  
The superiority of HilbertMamba and PeanoMamba becomes solid when addressing the two challenges in low-light image enhancement: 1) preserving fine-grained local interactions and 2) maintaining global dependency modeling. 

Their advantages include: 1) The continuity of the Hilbert and Peano scans ensures that subtle local image features, such as textures and edges, are sampled with high fidelity. This is particularly important in low-light conditions, where minor details significantly impact the perceived visual quality. 2)By providing a more uniform sampling of the spatial domain \(\Omega\), Hilbert Scan lowers the dispersion \(\varepsilon(P,\Omega)\). Under the assumption that the underlying image function \(f\) is H\"older continuous with exponent \(\alpha\), the worst-case error bound 
\[
E(P,\mathcal{F}) \le K'\,\varepsilon(P,\Omega)^\alpha,
\]
is directly minimized, resulting in higher fidelity in the reconstructed features.

In summary, the Hilbert and Peano scans, by their space-filling and fractal properties, offer mathematically rigorous strategies that significantly enhance the performance of Vision Mamba. 

We also introduce a spatial dispersion metric that quantifies discontinuities in scanning patterns by jointly considering the magnitude and frequency of index jumps. For full details and derivations, please refer to the supplementary material.

\begin{figure*}[h]
\setlength{\abovecaptionskip}{0.01cm} 
\setlength{\belowcaptionskip}{-0.3cm} 
\centering
\includegraphics[width=1\linewidth]{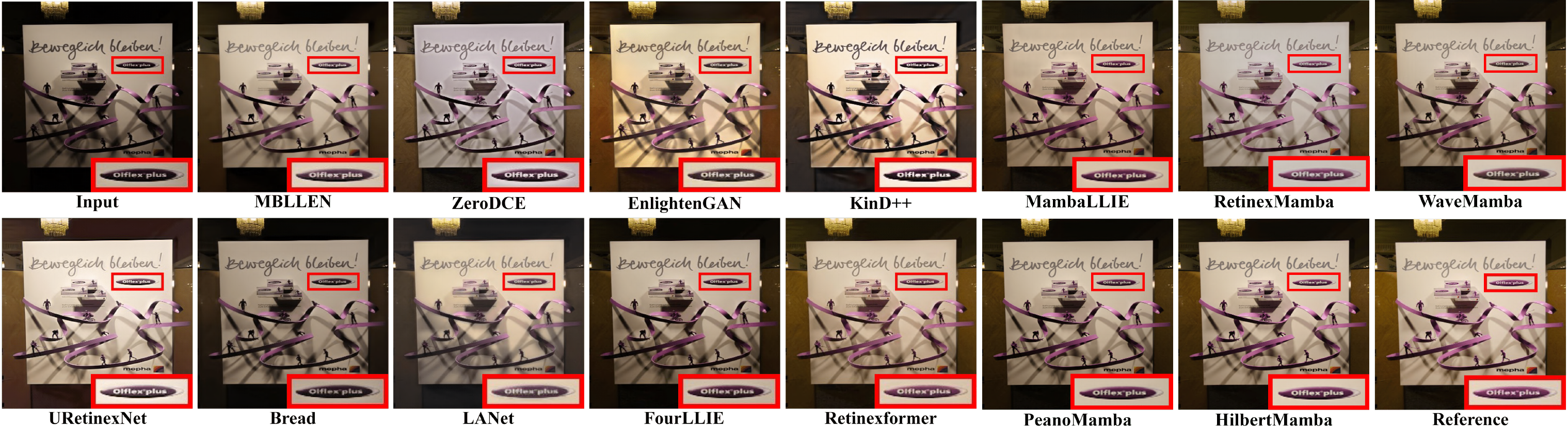}
\caption{\label{lolv2_syn}%
Visual comparisons of the enhanced results by different methods on LOLv2-synthetic.}
\end{figure*}

\section{Experiment}
\subsection{Experiment Settings}
\noindent \textbf{Datasets:}
We trained several Mamba-based low-light image enhancement methods \cite{bai2024retinexmamba, MambaLLIE, zou2024wavemamba} on two datasets: LOLv1 \cite{wei2018deep} and LOLv2 \cite{yang2021sparse}. LOLv1 contains 500 real-world low/normal-light image pairs, with 485 pairs for training and 15 for testing. LOLv2 is divided into LOLv2-real and LOLv2-synthetic subsets. LOLv2-real contains 689 paired images for training and 100 pairs for testing, collected by adjusting the exposure time and ISO. LOLv2-synthetic contains 900 paired images for training and 100 pairs for testing. 

\noindent \textbf{Evaluation metrics:}
To evaluate the performance of different methods and validate the effectiveness of the proposed method, we adopt full-reference image quality evaluation metrics to evaluate various low-light image enhancement approaches. We employ peak signal-to-noise ratio (PSNR), structural similarity (SSIM) \cite{wang2004image}, and learned perceptual image patch similarity (LPIPS) \cite{zhang2018unreasonable} as evaluation metrics to assess the model's performance. Higher PSNR and SSIM values and lower LPIPS scores generally indicate more significant similarity between two images.

\noindent \textbf{Methods Comparisons:} 
To demonstrate the superiority of the Mamba-based low-light image enhancement methods reinforced by scanning patterns with Hausdorff dimensions equal to 2, PeanoMamba and HilbertMamba, we compare these prompted baseline methods with a great variety of state-of-the-art methods both quantitatively and qualitatively, including RetinexNet \cite{wei2018deep}, MBLLEN \cite{lv2018mbllen}, ZeroDCE \cite{guo2020zero}, MIRNet \cite{zamir2020learning}, EnlightenGAN \cite{jiang2021enlightengan}, KinD++ \cite{zhang2021beyond}, LLFlow \cite{wang2022low}, URetinexNet \cite{wu2022uretinex}, SNRNet \cite{xu2022snr}, Bread \cite{guo2023low}, LANet \cite{yang2023learning}, FourLLIE \cite{wang2023fourllie}, Retinexformer \cite{cai2023retinexformer}. For fair comparison, all codes are downloaded from the authors' GitHub repositories, and all comparison results are gained from retraining based on their recommended experimental configurations.

\noindent \textbf{Scanning Paths Comparisons:} 
To demonstrate the superiority of the scanning patterns with Hausdorff dimensions equal to 2 in Mamba-based low-light image enhancement methods, we compare our proposed Hilbert Scan and Peano Scan with other mainstream types of scans: raster scan \cite{vim}, continuous scan \cite{zhou2024mambainmamba}, local scan \cite{LocalMamba}, and tree scan \cite{xiao2024grootvl}. We implement these scans on three different Mamba-based low-light image enhancement methods: RetinexMamba \cite{bai2024retinexmamba}, MambaLLIE \cite{MambaLLIE}, and WaveMamba \cite{zou2024wavemamba}, respectively, to show the generalizability of the Hilbert Scan and Peano Scan in Mamba-based low-light image enhancement methods. For fair comparison, all codes are downloaded from the authors' websites, and all comparison results are based on their recommended experimental configurations.

\noindent \textbf{Implementation details:}
We implement our HilbertMamba and PeanoMamba models using PyTorch and train it for 500000 iterations on an NVIDIA GeForce RTX 4090D GPU. The AdamW \cite{Adamw} optimizer ($\beta_{1} = 0.9, \beta_{2} = 0.99$) is adopted for optimization. initial learning rate $5\times 10^{-4}$  gradually reduced to $1\times 10^{-7}$ with the cosine annealing. During training, input images are cropped to 256×256 pixels to serve as training samples, with a batch size of 32. For the augmented training data, we use random rotations of 90, 180, 270, random flips, and random cropping to $256\times256$ size. To constrain the training of HilbertMamba and PeanoMamba, we use the $L_1$ loss function.

\begin{table*}[t]
\setlength{\abovecaptionskip}{0.1cm} 
\centering
\caption{Quantitative comparisons of different scans implemented on Mamba-based low-light image enhancement methods, Retinexmamba\cite{bai2024retinexmamba}, MambaLLIE\cite{MambaLLIE}, and WaveMamba\cite{zou2024wavemamba} on LOLv1, LOLv2-real, and LOLv2-synthetic. SS2D\cite{liu2024vmamba} represents raster scan, CS2D\cite{zhou2024mambainmamba} represents continuous scan, LS2D\cite{LocalMamba} represents local scan, TS2D\cite{xiao2024grootvl} represents tree scan, PS2D represents Peano scan and HS2D represents Hilbert scan.}
\begin{tabular}{c|c|c|ccc|ccc|ccc}
\toprule
\multicolumn{1}{c|}{\multirow{2}{*}{Methods}} & \multirow{2}{*}{Dimensions} & \multirow{2}{*}{Scans} & \multicolumn{3}{c|}{LOLv1} & \multicolumn{3}{c|}{LOLv2-real} & \multicolumn{3}{c}{LOLv2-synthetic} \\
\cmidrule(r){4-6} \cmidrule(r){7-9} \cmidrule(r){10-12}
 &  &  & PSNR$\uparrow$ & SSIM$\uparrow$ & LPIPS$\downarrow$ & PSNR$\uparrow$ & SSIM$\uparrow$ & LPIPS$\downarrow$ & PSNR$\uparrow$ & SSIM$\uparrow$ & LPIPS$\downarrow$ \\
\midrule
\multirow{6}{*}{RetinexMamba}
  & 1 & SS2D & 23.34 & \textbf{0.8328} & \textbf{0.0893} & 22.45 & 0.8444 & 0.1182 & 25.31 & 0.9291 & 0.0407 \\
  & 1 & CS2D & 23.43 & 0.8212 & 0.0967 & 26.75 & \underline{0.8656} & \underline{0.0768} & 25.27 & 0.9274 & 0.0428 \\
  & 1 & LS2D & 23.23 & 0.8165 & 0.0985 & 25.92 & 0.8581 & 0.0786 & 24.11 & 0.9116 & 0.0533 \\
  & 1 & TS2D & 22.74 & 0.8175 & 0.0991 & 25.97 & 0.8621 & 0.0793 & 25.34 & 0.9259 & 0.0447 \\
  & 2 & \textit{\underline{PS2D}} & \textbf{23.80} & \underline{0.8226} & \underline{0.0926} & \textbf{29.25} & \textbf{0.8765} & \textbf{0.0677} & \underline{25.54} & \textbf{0.9320} & \textbf{0.0382} \\
  & 2 & \textit{\underline{HS2D}} & \underline{23.64} & 0.8147 & 0.1034 & \underline{26.91} & 0.8614 & 0.0800 & \textbf{25.59} & \underline{0.9294} & \underline{0.0403} \\
\midrule
\multirow{6}{*}{MambaLLIE}
  & 1 & SS2D & 21.29 & 0.8236 & 0.0907 & 22.95 & 0.8466 & 0.1047 & 21.29 & 0.8236 & 0.0907 \\
  & 1 & CS2D & 21.96 & 0.8119 & 0.1011 & 26.17 & 0.8636 & 0.0811 & 23.59 & 0.9228 & 0.0462 \\
  & 1 & LS2D & 22.52 & 0.8197 & 0.0970 & 28.85 & 0.8760 & 0.0721 & 23.42 & 0.9191 & 0.0489 \\
  & 1 & TS2D & 18.79 & 0.7867 & 0.1329 & 21.32 & 0.8349 & 0.1080 & 22.37 & 0.9038 & 0.0648 \\
  & 2 & \textit{\underline{PS2D}} & \underline{22.62} & \underline{0.8275} & \underline{0.0899} & \underline{29.76} & \textbf{0.8874} & \textbf{0.0647} & \underline{24.74} & \underline{0.9316} & \textbf{0.0385} \\
  & 2 & \textit{\underline{HS2D}} & \textbf{23.36} & \textbf{0.8344} & \textbf{0.0870} & \textbf{29.99} & \underline{0.8863} & \underline{0.0668} & \textbf{24.90} & \textbf{0.9324} & \underline{0.0398} \\
\midrule
\multirow{6}{*}{WaveMamba}
  & 1 & SS2D & 23.36 & 0.8544 & 0.1164 & 29.04 & 0.9080 & 0.0895 & 24.63 & 0.9227 & 0.0737 \\
  & 1 & CS2D & 23.40 & 0.8515 & 0.1223 & 28.83 & 0.9080 & 0.0933 & 24.29 & 0.9212 & 0.0711 \\
  & 1 & LS2D & 23.92 & 0.8600 & \underline{0.1123} & 29.13 & 0.9048 & 0.0879 & 24.49 & 0.9260 & 0.0640 \\
  & 1 & TS2D & 23.79 & 0.8568 & 0.1199 & 28.39 & 0.9009 & 0.0922 & 24.92 & 0.9293 & 0.0615 \\
  & 2 & \textit{\underline{PS2D}} & \underline{23.99} & \underline{0.8571} & \textbf{0.1097} & \underline{30.71} & \textbf{0.9175} & \textbf{0.0778} & \underline{25.32} & \textbf{0.9382} & \textbf{0.0458} \\
  & 2 & \textit{\underline{HS2D}} & \textbf{24.85} & \textbf{0.8617} & 0.1173 & \textbf{31.04} & \underline{0.9148} & \underline{0.0855} & \textbf{25.74} & \underline{0.9369} & \underline{0.0498} \\
\bottomrule
\end{tabular}
\label{tab2}
\end{table*}

\subsection{Comparison with State-of-the-Art Methods}

\noindent \textbf{Results on paired datasets:}
Table~\ref{tab1} displays the quantitative results obtained from comparison methods, where it can be observed that our proposed methods outperform others in most cases, nearly securing the second-best results where they fall short. 
When compared with the recent transformer-based approaches, SNR \cite{xu2022snr} and Retinexformer \cite{cai2023retinexformer}, our method achieves 1.11, 5.94, and 3.10 dB improvement on LOLv1, LOLv2-real, and LOLv2-synthetic datasets. Especially on LOLv2-real, the improvement is over 5 dB, as shown in Table~\ref{tab1}. 
Compared with the recent Fourier-based method, FourLLIE \cite{wang2023fourllie}, our WaveletMamba yields 5.23, 6.38, and 4.12 dB on the three benchmarks. 
When compared with the recent CNN-based approaches, Bread \cite{guo2023low} and LANet \cite{yang2023learning}, our WaveletMamba gains 4.52, 3.43, and 11.78 dB on the three datasets in Table~\ref{tab1}. 
Particularly, our method yields the best visually appealing results in real-world images, as shown in Fig.~\ref{lolv2_real} and Fig.~\ref{lolv2_syn}, where our method effectively suppresses noise and restores image details, resulting in visuals that closely resemble the original scene. Please zoom in for a better view. The visual results on LOLv1 are provided in the supplementary material. All these results suggest the outstanding effectiveness and efficiency advantage of our HilbertMamba and PeanoMamba.

\subsection{Comparison with Different Scanning Paths}
\noindent \textbf{Results with Different Scanning Patterns:}
Table~\ref{tab2} summarizes the performance of RetinexMamba \cite{bai2024retinexmamba}, MambaLLIE \cite{MambaLLIE}, and WaveMamba \cite{zou2024wavemamba} on the LOLv1, LOLv2-real, and LOLv2-synthetic datasets using various 2D scanning paths. Our results clearly show that the dimension-2 scans, PS2D (Peano) and HS2D (Hilbert), consistently outperform the dimension-1 scans (SS2D, CS2D, LS2D, TS2D) in terms of PSNR, SSIM, and LPIPS.

\begin{figure}[t]
\setlength{\abovecaptionskip}{0.01cm} 
\setlength{\belowcaptionskip}{-0.3cm} 
\centering
\includegraphics[width=1\linewidth]{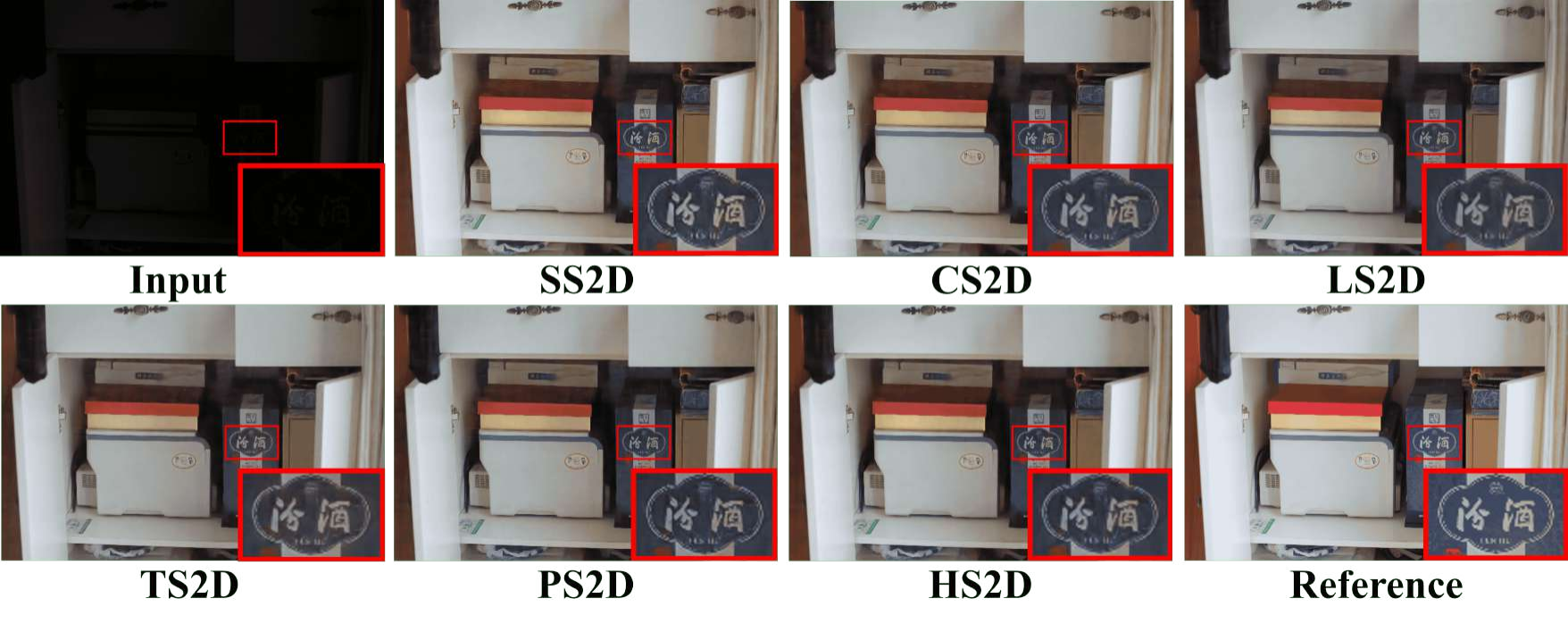}
\caption{\label{fig1}%
Visual comparisons of results of different scanning paths (based on WaveMamba\cite{zou2024wavemamba}) on LOLv1.}
\end{figure}

For example, RetinexMamba with PS2D yields PSNRs of 23.8 dB (LOLv1), 29.25 dB (LOLv2-real), and 25.54 dB (LOLv2-synthetic). Similarly, MambaLLIE achieves PSNRs of 22.62 dB and 23.36 dB on LOLv1 and 29.76 dB and 29.99 dB on LOLv2-real using PS2D and HS2D, respectively. Notably, WaveMamba with HS2D reaches the highest PSNRs of 24.85 dB on LOLv1 and 31.04 dB on LOLv2-real, surpassing its performance with any dimension-1 scan. The qualitative comparisons implemented on the method WaveMamba \cite{zou2024wavemamba} are shown in Fig~\ref{fig1}, Fig~\ref{fig2} and Fig~\ref{fig3}. The qualitative comparisons of the other two methods are provided in the supplementary material.

The dimension-1 scans include the normal raster selective scan (SS2D) \cite{vim}, continuous scan (CS2D) \cite{zhou2024mambainmamba}, local scan (LS2D) \cite{LocalMamba}, and tree scan (TS2D) \cite{xiao2024grootvl}. Visualizations for SS2D, CS2D, and LS2D are provided in the supplementary material, while TS2D, being input-dependent, is detailed in \cite{xiao2024grootvl}.

\begin{figure}[t]
\setlength{\abovecaptionskip}{0.01cm} 
\setlength{\belowcaptionskip}{-0.3cm} 
\centering
\includegraphics[width=1\linewidth]{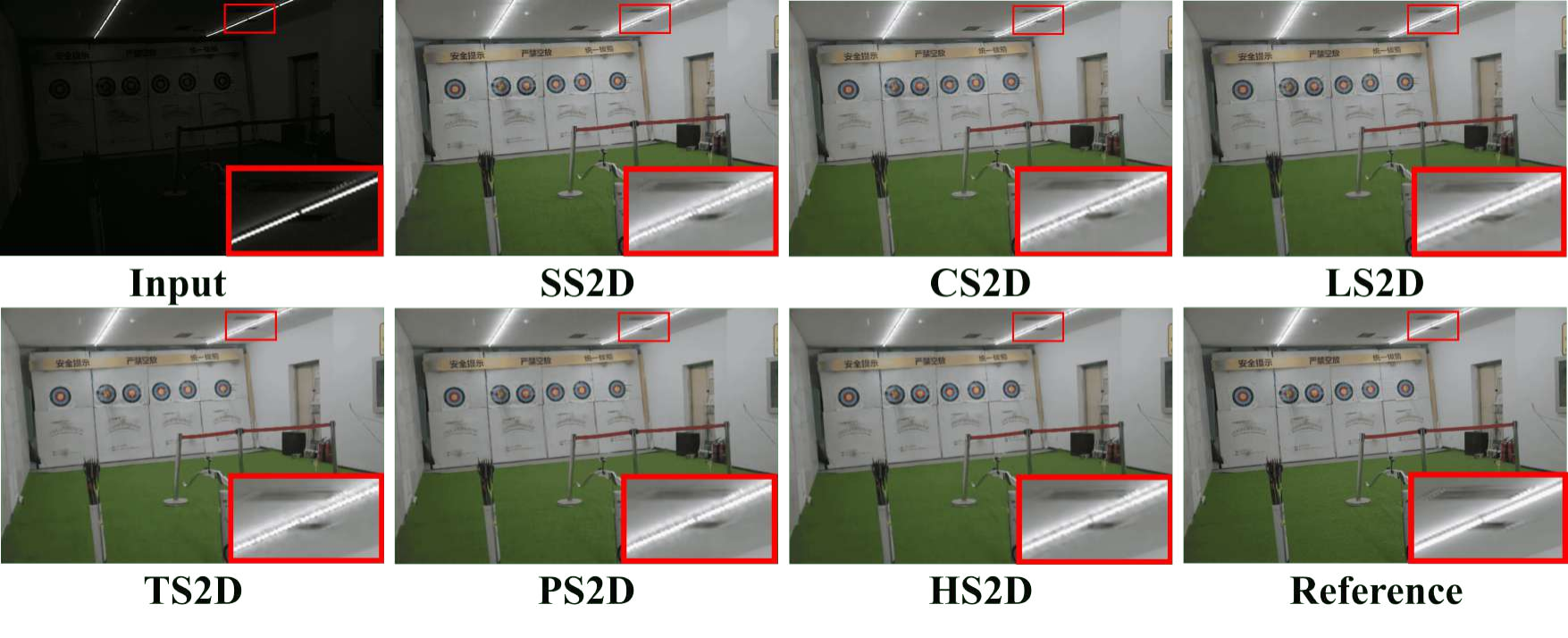}
\caption{\label{fig2}%
Visual comparisons of results of different scanning paths (based on WaveMamba\cite{zou2024wavemamba}) on LOLv2-real.}
\end{figure}

\begin{figure}[h]
\setlength{\abovecaptionskip}{0.01cm} 
\setlength{\belowcaptionskip}{-0.3cm} 
\centering
\includegraphics[width=0.85\linewidth]{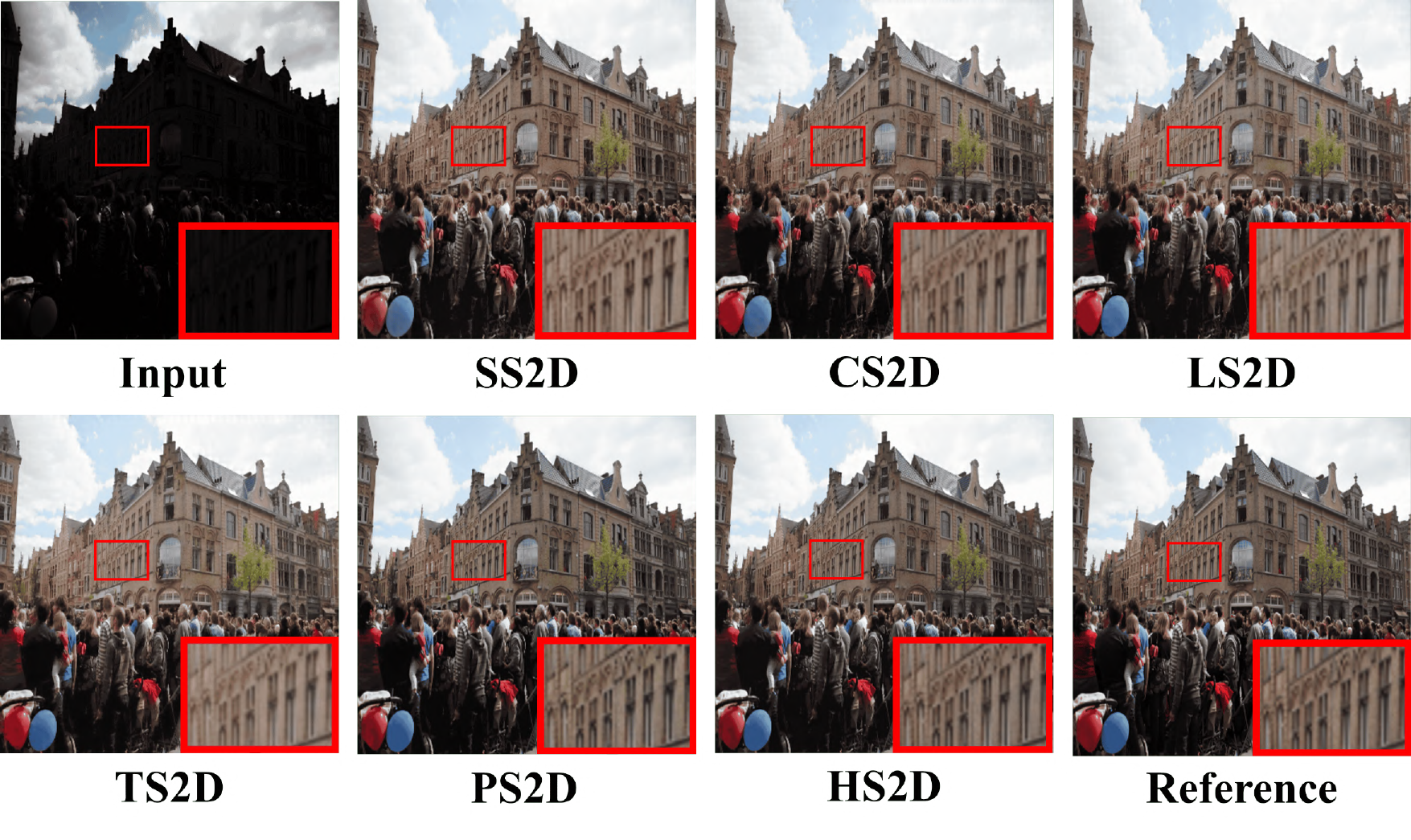}
\caption{\label{fig3}%
Visual comparisons of results of different scanning paths (based on WaveMamba\cite{zou2024wavemamba}) on LOLv2-synthetic. }
\end{figure}

\section{Conclusion}
In this paper, we rigorously demonstrate both mathematically and empirically that the performance of Mamba‐based low-light image enhancement methods can be significantly improved by adopting scanning patterns with higher Hausdorff dimensions. By introducing HilbertMamba and PeanoMamba, our work pioneers a novel scanning paradigm that transforms the way state space models (SSMs) handle two-dimensional image data. Leveraging the fractal and space‐filling properties of Hilbert and Peano curves, our approach reorders image pixels into a one-dimensional sequence in a manner that preserves delicate local features and enhances the reconstruction of fine-scale details under challenging low-light conditions. Our methodology effectively overcomes the inherent limitations of conventional raster scanning by reducing the approximation errors typical in traditional mapping strategies. This allows SSMs to capture long-range dependencies while faithfully reconstructing local textures and edges critical for perceptual image quality. Extensive experiments on standard low-light benchmarks validate the theoretical benefits of high Hausdorff dimension scanning patterns, with our methods consistently achieving superior quantitative metrics and visually pleasing enhancement results compared to state-of-the-art approaches. In summary, by bridging the gap between fractal geometry and state space modeling, HilbertMamba and PeanoMamba not only advance the state-of-the-art in low-light image enhancement but also offer a versatile framework that could reshape how we approach sampling and representation in a wide range of vision applications.

\bibliographystyle{ACM-Reference-Format}
\bibliography{main}

\end{document}